\def\year{2019}\relax
\documentclass[letterpaper]{article} 
\usepackage{aaai19}  
\usepackage{times}  
\usepackage{helvet}  
\usepackage{courier}  
\usepackage{url}  
\usepackage{graphicx}  

\usepackage[usenames, dvipsnames]{xcolor}
\usepackage{multirow}
\usepackage{algorithm}
\usepackage{algorithmic}
\usepackage{amsfonts}
\usepackage{mathtools,amssymb}
\usepackage{fixltx2e}
\usepackage{bbold}
\usepackage{url}

\usepackage{array}

\usepackage{setspace}

\newcommand{\eqn}[1]{Equation~#1}
\newcommand{\fig}[1]{Fig.~#1}
\newcommand{\tab}[1]{Table~#1}

\usepackage{expl3}
\ExplSyntaxOn
\newcommand\latinabbrev[1]{
	\peek_meaning:NTF . {
		#1\@}%
	{ \peek_catcode:NTF a {
			#1.\@ }%
		{#1.\@}}}
\ExplSyntaxOff
\def\eg{\latinabbrev{e.g}}
\def\etal{\latinabbrev{et al}}

\def\ie{\latinabbrev{i.e}}

\begin{document}
%
\title{MonoGRNet: A Geometric Reasoning Network for \\Monocular 3D Object Localization}
\author{
Zengyi Qin$^{1, 2}$ \quad	Jinglu Wang$^{3}$ \quad  \quad Yan Lu$^{3}$\\
$^{1}$Tsinghua University \quad $^{2}$Massachusetts Institute of Technology \quad $^{3}$Microsoft Research \\
qinzy@mit.edu \quad jinglwa@microsoft.com \quad yanlu@microsoft.com \\
}

\maketitle

\begin{abstract}

Localizing objects in the real 3D space, which plays a crucial role in scene understanding, is particularly challenging given only a single RGB image due to the geometric information loss during imagery projection. We propose MonoGRNet for monocular 3D object detection and localization from a single RGB image via geometric reasoning in both the observed 2D projection and the unobserved depth dimension. MonoGRNet is a single, unified network composed of four task-specific subnetworks, responsible for 2D object detection, instance depth estimation (IDE), 3D localization and local corner regression. Unlike the pixel-level depth estimation that needs per-pixel annotations, we propose a novel IDE method that directly predicts the depth of the targeting 3D bounding box's center using sparse supervision. The 3D localization is further achieved by estimating the position in the horizontal and vertical dimensions. Finally, MonoGRNet is jointly learned by optimizing the locations and poses of the 3D bounding boxes in the global context. We demonstrate that MonoGRNet achieves state-of-the-art performance on challenging datasets. See the project website \url{https://sites.google.com/view/monogrnet}\end{abstract}
\section{Introduction}
Typical object localization or detection from a RGB image estimates 2D boxes that bound visible parts of the objects belonging to the specific classes on the image plane. However, this kind of result cannot provide geometric perception in the real 3D world for scene understanding, which is crucial for applications, such as robotics, mixed reality, and autonomous driving.

In this paper, we address the problem of localizing amodal 3D bounding boxes (ABBox-3D) of objects at their full extents from a monocular RGB image. Unlike 2D analysis on the image plane, 3D localization with the extension to an unobserved dimension, \ie, depth, not solely enlarges the searching space but also introduces inherent ambiguity of 2D-to-3D mapping, increasing the task's difficulty significantly.

\begin{figure*}
	\centering 
	\includegraphics[width=1\linewidth,trim={0 0cm 0cm 0cm},clip]{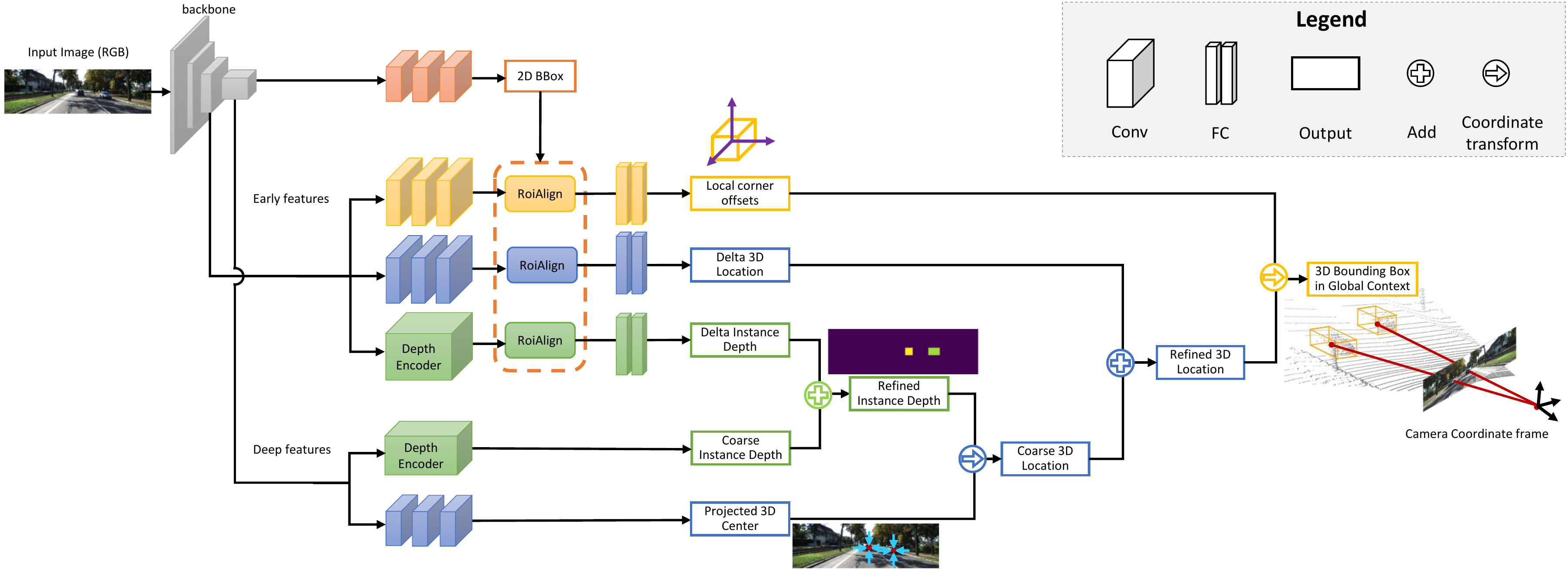}
	\caption{\textbf{MonoGRNet for 3D object localization from a monocular RGB image.} MonoGRNet consists of four subnetworks for 2D detection(\textcolor[RGB]{237,125,49}{brown}), instance depth estimation(\textcolor[RGB]{112,173,71}{green}), 3D location estimation(\textcolor[RGB]{91,155,213}{blue}) and local corner regression(\textcolor[RGB]{255,192,0}{yellow}). Guided by the detected 2D bounding box, the network first estimates depth and 2D projection of the 3D box's center to obtain the global 3D location, and then regresses corner coordinates in local context. The final 3D bounding box is optimized in an end-to-end manner in the global context based on the estimated 3D location and local corners. (Best viewed in color.)}
	\label{fig:net}  
\end{figure*}

Most state-of-the-art monocular methods~\cite{xu2018multifusion,zhuo20183dbp} estimate pixel-level depths and then regress 3D bounding boxes. Nevertheless, pixel-level depth estimation does not focus on object localization by design. It aims to minimize the mean error for all pixels to get an average optimal estimation over the whole image, while objects covering small regions are often neglected~\cite{fu2018ordinal}, which drastically downgrades the 3D detection accuracy.
    
We propose MonoGRNet, a unified network for amodal 3D object localization from a monocular image. Our key idea is to decouple the 3D localization problem into several progressive sub-tasks that are solvable using only monocular RGB data. The network starts from perceiving semantics in 2D image planes and then performs geometric reasoning in the 3D space. 

A challenging problem we overcome is to accurately estimate the depth of an instance's 3D center without computing pixel-level depth maps. We propose a novel instance-level depth estimation (IDE) module, which explores large receptive fields of deep feature maps to capture coarse instance depths and then combines early features of a higher resolution to refine the IDE.

To simultaneously retrieve the horizontal and vertical position, we first predict the 2D projection of the 3D center. In combination with the IDE, we then extrude the projected center into real 3D space to obtain the eventual 3D object location. All the components are integrated into the end-to-end network, MonoGRNet, featuring its three 3D reasoning branches illustrated in \fig{\ref{fig:net}}, and is finally optimized by a joint geometric loss that minimizes the 3D bounding box discrepancy in the global context.

We argue that RGB information alone can provide almost accurate 3D locations and poses of objects. Experiments on the challenging KITTI dataset demonstrate that our network outperforms the state-of-art monocular method in 3D object localization with the least inference time.
In summary, our contributions are three-fold:
\begin{itemize}
	
	\item A novel instance-level depth estimation approach that directly predicts central depths of ABBox-3D in the absence of dense depth data, regardless of object occlusion and truncation.
	\item A progressive 3D localization scheme that explores rich feature representations in 2D images and extends geometric reasoning into 3D context.
	\item A unified network that coordinates localization of objects in 2D, 2.5D and 3D spaces via a joint optimization, which performs efficient inference (taking $\sim$0.06s/image).

\end{itemize}

\section{Related Work}
Our work is related to 3D object detection and monocular depth estimation. We mainly focus on the works of studying 3D detection and depth estimation, while 2D detection is the basis for coherence.

\paragraph{2D Object Detection.}
2D object detection deep networks are extensively studied.
Region proposal based methods~\cite{girshick2015fast,ren2017faster} generate impressive results but perform slowly due to complex multi-stage pipelines. 
Another group of methods~\cite{redmon2016yolo,redmon2017yolo9000,liu2016ssd,fu2017dssd} focusing on fast training and inferencing apply a single stage detection.
Multi-net~\cite{teichmann2016multinet} introduces an encoder-decoder architecture for real-time semantic reasoning. Its detection decoder combines the fast regression in Yolo~\cite{redmon2016yolo} with the size-adjusting RoiAlign of Mask-RCNN~\cite{he2017mrcn}, achieving a satisfied speed-accuracy ratio.
All these methods predict 2D bounding boxes of objects while none 3D geometric features are considered.
\paragraph{3D Object Detection.}
Existing methods range from single-view RGB~\cite{chen2016monocular,xu2018multifusion,chabot2017deepmanta,kehl2017ssd6d}, multi-view RGB~\cite{chen2017multiview,chen20153dop,wang2016image}, to RGB-D~\cite{qi2017frustum,song2016deep,liu2015higher,zhang2014joint}.
While geometric information of the depth dimension is provided, the 3D detection task is much easier.
Given RGB-D data, FPointNet~\cite{qi2017frustum} extrudes 2D region proposals to a 3D viewing frustum and then segments out the point cloud of interest object.
MV3D~\cite{chen2017multiview} generates 3D object proposals from bird's eye view maps given LIDAR point clouds, and then fuses features in RGB images, LIDAR front views and bird's eye views to predict 3D boxes.
3DOP\cite{chen20153dop} exploits stereo information and contextual models specific to autonomous driving. 

The most related approaches to ours are using a monocular RGB image. Information loss in the depth dimension significantly increases the task's difficulty. Performances of state-of-the-art such methods still have large margins to RGB-D and multi-view methods. 
Mono3D \cite{chen2016monocular} exploits segmentation and context priors to generate 3D proposals. Extra networks for semantic and instance segmentation are required, which cost more time for both training and inference. 
Xu \etal \cite{xu2018multifusion} leverage a pretrained disparity estimation model~\cite{mahjourian2018unsupervised} to guide the geometry reasoning. 
Other methods~\cite{chabot2017deepmanta,kehl2017ssd6d} utilize 3D CAD models to generate synthetic data for training, which provides 3D object templates, object poses and their corresponding 2D projections for better supervision.
All previous methods exploit additional data and networks to facilitate the 3D perception, while our method only requires annotated 3D bounding boxes and no extra network needs to train. This makes our network much light weighted and efficient for training and inference.

\paragraph{Monocular Depth Estimation.}
Recently, although many pixel-level depth estimation networks~\cite{fu2018ordinal,eigen2015predicting} have been proposed, they are not sufficient for 3D object localization.
When regressing the pixel-level depth, the loss function takes into account every pixel in the depth map and treats them without significant difference. In a common practice, the loss values from each pixel are summed up as a whole to be optimized. Nevertheless, there is a likelihood that the pixels lying in an object are much fewer than those lying in the background, and thus the low average error does not indicate the depth values are accurate in pixels contained in an object. In addition, dense depths are often estimated from disparity maps that may produce large errors at far regions, which may downgrade the 3D localization performance drastically.

Different from the abovementioned pixel-level depth estimation methods, we are the first to propose an instance-level depth estimation network which jointly takes semantic and geometric features into account with sparse supervision data. 
\section{Approach}
\begin{figure}
	\centering
	\includegraphics[width=1\linewidth]{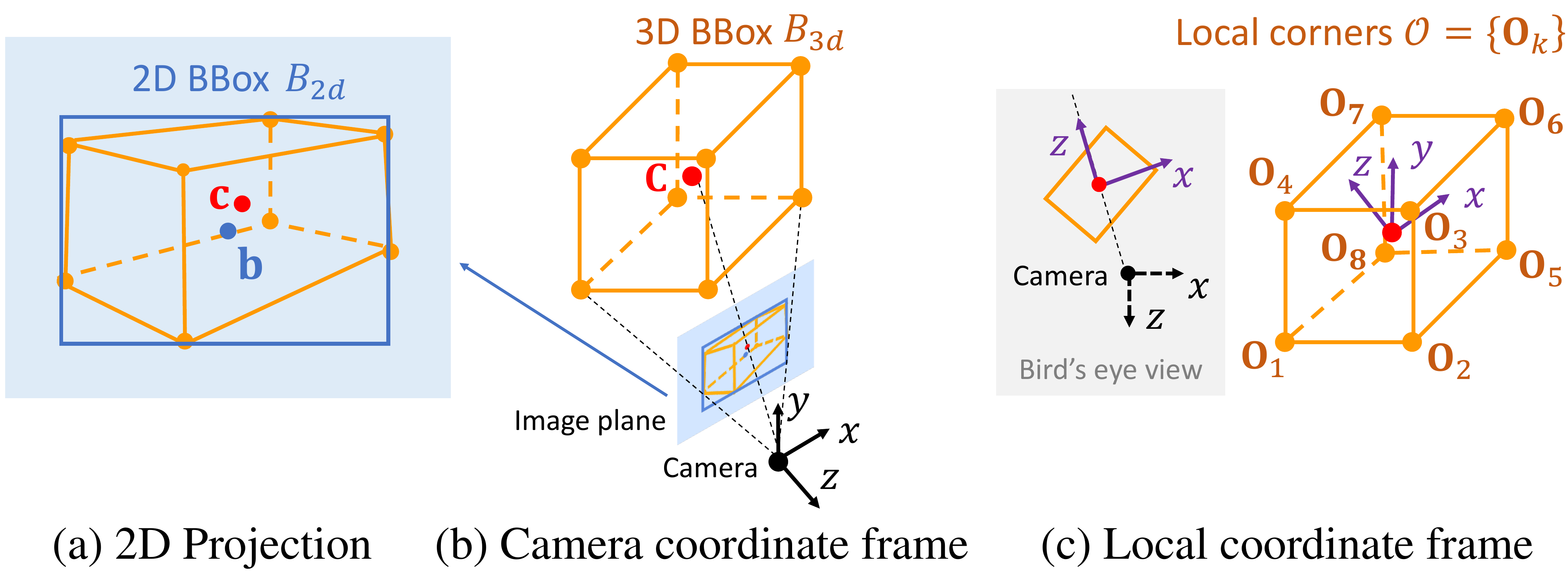}
	\caption{Notation for 3D bounding box localization.}
	\label{fig:notation}
\end{figure}

We propose an end-to-end network, MonoGRNet, that directly predicts ABBox-3D from a single RGB image. MonoGRNet is composed of a 2D detection module and three geometric reasoning subnetworks for IDE, 3D localization and ABBox-3D regression. In this section, we first formally define the 3D localization problem and then detail MonoGRNet for four subnetworks.

\subsection{Problem Definition}
Given a monocular RGB image, the objective is to localize objects of specific classes in the 3D space. A target object is represented by a class label and an ABBox-3D, which bounds the complete object regardless of occlusion or truncation. An ABBox-3D is defined by a 3D center $\mathbf{C}=(X_c,Y_c,Z_c)$ in global context and eight corners $\mathcal{O}=\{\mathbf{O}_k\}, k=1,...,8$, related to local context. The 3D location $\mathbf{C}$ is calibrated in the camera coordinate frame and the local corners $\mathcal{O}$ are in a local coordinate frame, shown in \fig{\ref{fig:notation}} (b) and (c) respectively.

We propose to separate the 3D localization task into four progressive sub-tasks that are resolvable using only a monocular image. First, the 2D box $B_{2d}$ with a center $\mathbf{b}$ and size $(w,h)$ bounding the projection of the ABBox-3D is detected. Then, the 3D center $\mathbf{C}$ is localized by predicting its depth $Z_c$ and 2D projection $\mathbf{c}$. Notations are illustrated in \fig{\ref{fig:notation}}.
Finally, local corners $\mathcal{O}$ with respect to the 3D center are regressed based on local features.
In summary, we formulate the ABBox-3D localization as estimating the following parameters of each interest object:
\begin{eqnarray}
\label{eq:abbox-3d}
B_{3d}=(B_{2d}, Z_c, \mathbf{c}, \mathcal{O})
\end{eqnarray}

\subsection{Monocular Geometric Reasoning Network}
MonoGRNet is designed to estimate four components, $B_{2d}$, $Z_c$, $\mathbf{c}$, $\mathcal{O}$, with four subnetworks respectively. Following a CNN backbone, they are integrated into a unified framework, as shown in \fig{\ref{fig:net}}.
\subsubsection{2D Detection.}
The 2D detection module is the basic module that stabilizes the feature learning and also reveals regions of interest to the subsequent geometric reasoning modules.

We leverage the design of the detection component in~\cite{teichmann2016multinet}, which combines fast regression~\cite{redmon2016yolo} and size-adaptive RoiAlign~\cite{he2017mrcn}, to achieve a convincing speed-accuracy ratio.
An input image $I$ of size $W \times H$ is divided into an $S_x \times S_y$ grid $\mathcal{G}$, where a cell is indicated by $\mathbf{g}$. The output feature map of the backbone is also reduced to $S_x \times S_y$. Each pixel in the feature map corresponding to an image grid cell yields a prediction. The 2D prediction of each cell $\mathbf{g}$ contains the confidence that an object of interest is present and the 2D bounding box of this object, namely, $(Pr_{obj}^{\mathbf{g}}, B_{2d}^{\mathbf{g}})$, indicated by a superscript $\mathbf{g}$. The 2D bounding box $B_{2d}=(\delta_{x_b},\delta_{y_b},w,h)$ is represented by the offsets $(\delta_{x_b}, \delta_{y_b})$ of its center $\mathbf{b}$ to the cell $\mathbf{g}$ and the 2D box size $(w,h)$. 

The predicted 2D bounding boxes are taken as inputs of the RoiAlign~\cite{he2017mrcn} layers to extract early features with high resolutions to refine the predictions and reduce the performance gap between this fast detector with proposal-based detectors.

\subsubsection{Instance-Level Depth Estimation.}
\begin{figure}[t]
	\centering
	\includegraphics[width=1\linewidth]{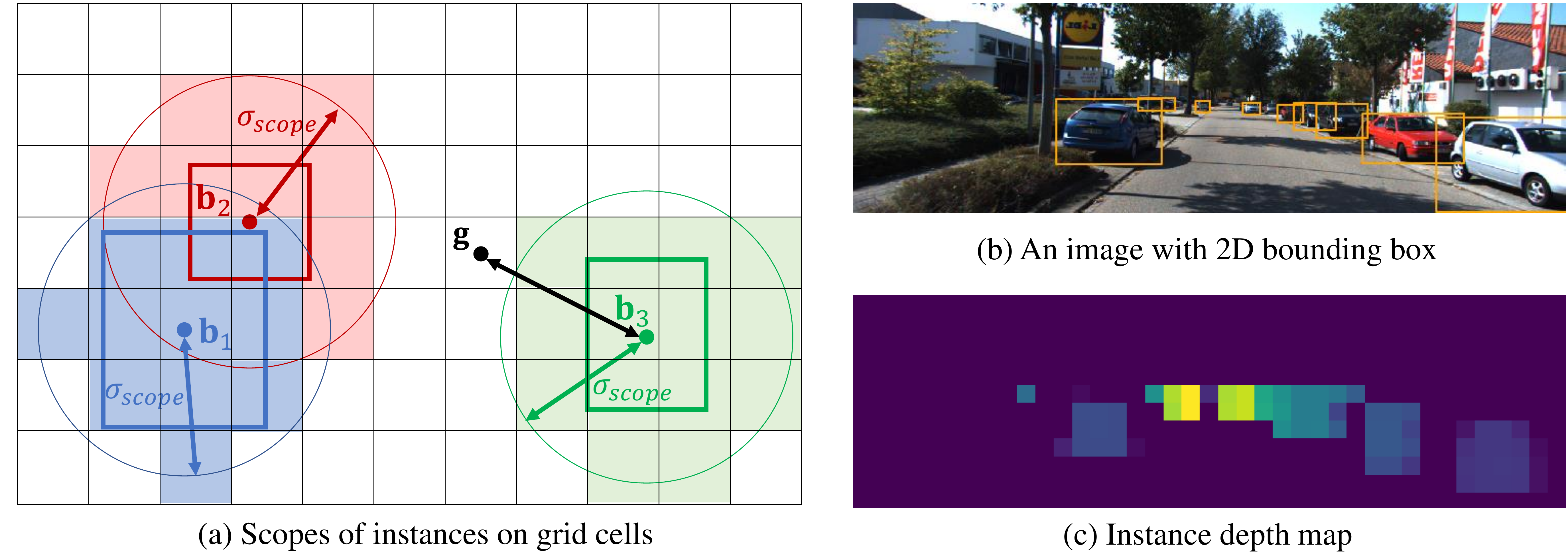}
	\caption{Instance depth. (a) Each grid cell $\mathbf{g}$ is assigned to a nearest object within a distance $\sigma_{scope}$ to the 2D bbox center $\mathbf{b}_i$. Objects closer to the camera are assigned to handle occlusion. Here $Z_c^1 < Z_c^2$. (b) An image with detected 2D bounding boxes. (c) Predicted instance depth for each cell.}
	\label{fig:instance_depth}
\end{figure}

The IDE subnetwork estimates the depth of the ABBox-3D center $Z_c$. Given the divided grid $\mathcal{G}$ in the feature map from backbone, each grid cell $\mathbf{g}$ predicts the 3D central depth of the nearest instance within a distance threshold $\sigma_{scope}$, considering depth information, \ie, closer instances are assigned for cells, as illustrated in \fig{\ref{fig:instance_depth}} (a). An example of predicted instance depth for each cell is shown in \fig{\ref{fig:instance_depth}} (c).

\begin{figure}[t]
	\centering
	\includegraphics[width=1\linewidth]{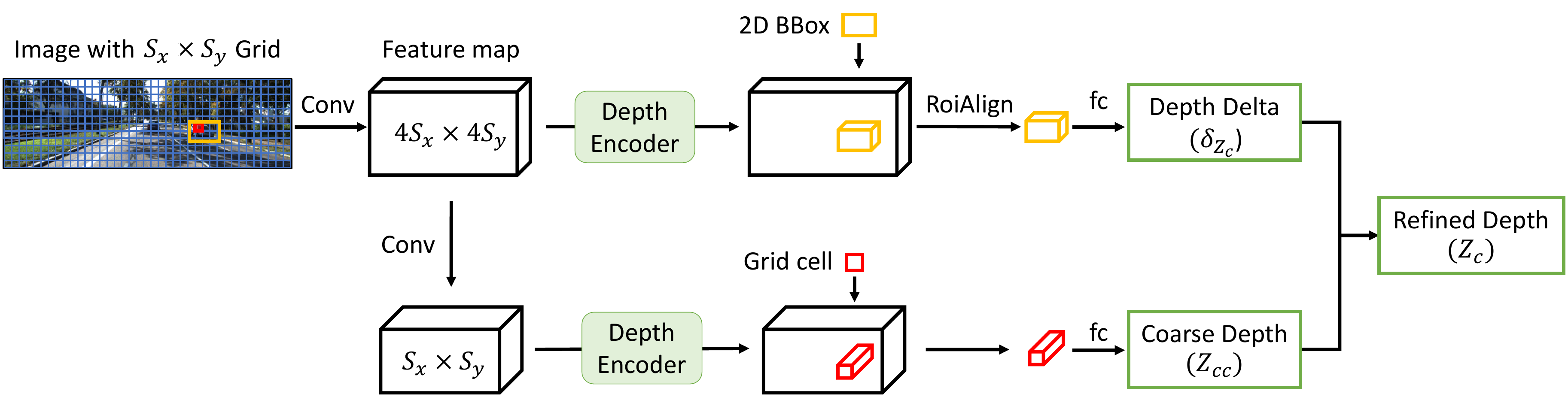}
	\caption{Instance depth estimation subnet. This shows the inference of the red cell.}
	\label{fig:instancedepth_net}
\end{figure}

The IDE module consists of a coarse regression of the region depth regardless of the scales and specific 2D location of the object, and a refinement stage that depends on the 2D bounding box to extract encoded depth features at exactly the region occupied by the target, as illustrated in \fig{\ref{fig:instancedepth_net}}.

Grid cells in deep feature maps from the CNN backbone have larger receptive fields and lower resolution in comparison with that of shallow layers. Because they are less sensitive to the exact location of the targeted object, it is reasonable to regress a coarse depth offset $Z_{cc}$ from deep layers. Given the detected 2D bounding box, we are able to perform RoiAlign to the region containing an instance in early feature maps with a higher resolution and a smaller receptive field. The aligned features are passed through fully connected layers to regress a delta $\delta_{Z_c}$ in order to refine the instance-level depth value. The final prediction is $Z_c = Z_{cc} +\delta_{Z_c}$. 
\subsubsection{3D Location Estimation.}
This subnetwork estimates the location of 3D center $\mathbf{C}=(X_c,Y_c,Z_c)$ of an object of interest in each grid $\mathbf{g}$. As illustrated in \fig{\ref{fig:notation}}, the 2D center $\mathbf{b}$ and the 2D projection $\mathbf{c}$ of $\mathbf{C}$ are not located at the same position due to perspective transformation. We first regress the projection $\mathbf{c}$ and then backproject it to the 3D space based on the estimated depth $Z_c$.


In a calibrated image, we elaborate the projection mapping from a 3D point $\mathbf{X}=(X,Y,Z)$ to a 2D point $\mathbf{x}=(u,v)$, $\psi_{3D \mapsto 2D}: \mathbf{X} \mapsto \mathbf{x} $, by
\begin{align}
\label{eq:3d_2d}
u=f_x * X / Z + p_x, & \quad &
v=f_y * Y / Z + p_y
\end{align}
where $f_x$ and $f_y$ are the focal length along X and Y axes, $p_x$ and $p_y$ are coordinates of the principle point.
Given known $Z$, the backprojection mapping $\psi_{2D \mapsto 3D}: (\mathbf{x},Z) \mapsto \mathbf{X} $ takes the form:
\begin{align}
\label{eq:2d_3d}
X=(u-p_x) * Z / f_x, & \quad &
Y=(v-p_y) * Z / f_y
\end{align}

Since we have obtained the instance depth $Z_c$ from the IDE module, the 3D location $\mathbf{C}$ can be analytically computed using the 2D projected center $\mathbf{c}$ according to \eqn{\ref{eq:2d_3d}}. Consequently, the 3D estimation problem is converted to a 2D keypoint localization task that only relies on a monocular image. 

Similar to the IDE module, we utilize deep features to regress the offsets $\delta_{\mathbf{c}}=(\delta_{x_c},\delta_{y_c})$ of a projected center $\mathbf{c}$ to the grid cell $\mathbf{g}$ and calculate a coarse 3D location $\mathbf{C}_s = \psi_{2D \mapsto 3D}(\delta_{\mathbf{c}}+\mathbf{g},Z_c)$. In addition, the early features with high resolution are extracted to regress the delta $\delta_{\mathbf{C}}$ between the predicted $\mathbf{C}_s$ and the groundtruth $\widetilde{\mathbf{C}}$ to refine the final 3D location, $\mathbf{C}=\mathbf{C}_s+\delta_{\mathbf{C}}$.

\subsection{3D Box Corner Regression}


This subnetwork regresses eight corners, \ie, $\mathcal{O}=\{\mathbf{O}_k\}, k=1,...,8$, in a local coordinate frame. Since each grid cell predicts a 2D bounding box in the 2D detector, we apply RoiAlign to the cell's corresponding region in early feature maps with high resolution and regress the local corners of the 3D bounding box. 

In addition, regressing poses of 3D boxes in camera coordinate frame is ambiguous ~\cite{xu2018shape,cao2018shape2,vitor2018shape3}. Even the poses of two 3D boxes are different, their projections could be similar when observed from certain viewpoints. We are inspired by Deep3DBox~\cite{mousavian20173dbox}, which regresses boxes in a local system according to the observation angle.

We construct a local coordinate frame, where the origin is at the object's center, the z-axis points straight from the camera to the center in bird's eye view, the x-axis is on the right of z-axis, and the y-axis does not change, illustrated in \fig{\ref{fig:notation}} (c). The transformation from local coordinates to camera coordinates are involved with a rotation $\mathbf{R}$ and a translation $\mathbf{C}$, and we obtain $\mathbf{O}_k^{cam}=\mathbf{R}\mathbf{O}_k + \mathbf{C}$, where $\mathbf{O}_k^{cam}$ are the global corner coordinates.
This is a single mapping between the perceived and actual rotation, in order to avoid confusing the regression model.  
\subsection{Loss Functions}
Here we formally formulate four task losses for the above subnetworks and a joint loss for the unified network. All the predictions are modified with a superscript $\mathbf{g}$ for the corresponding grid cell $\mathbf{g}$. Groundtruth observations are modified by the $\widetilde{(\cdot)}$ symbol.
\paragraph{2D Detection Loss.}
The object confidence is trained using softmax $(s\cdot)$ cross entropy $(CE)$ loss and the 2D bounding boxes $B_{2d}=(x_b,y_b,w,h)$ are regressed using a masked L1 distance loss. Note that $w$ and $h$ are normalized by $W$ and $H$. The 2D detection loss are defined as:
\begin{align}
&\mathcal{L}_{conf} = CE_{^{\mathbf{g}} \in \mathcal{G}}(s\cdot(Pr_{obj}^{\mathbf{g}}), \widetilde{Pr}_{obj}^{\mathbf{g}}) \nonumber\\
&\mathcal{L}_{bbox} = \sum_{\mathbf{g}} \mathbb{1}_{\mathbf{g}}^{obj}\cdot d(B_{2d}^{\mathbf{g}} , \widetilde{B}_{2d}^{\mathbf{g}}) \nonumber \\
&\mathcal{L}_{2d} = \mathcal{L}_{conf} + \omega \mathcal{L}_{bbox}
\end{align}
where $Pr_{obj}$ and $\widetilde{Pr}_{obj}$ refer to the confidence of predictions and groundtruths respectively, $d(\cdot)$ refers to L1 distance and $\mathbb{1}_{\mathbf{g}}^{obj}$ masks off the grids that are not assigned any object. The mask function $\mathbb{1}_{\mathbf{g}}^{obj}$ for each grid $\mathbf{g}$ is set to 1 if the distance between $\mathbf{g}$ $\mathbf{b}$ is less then $\sigma_{scope}$, and 0 otherwise.
The two components are balanced by $\omega$.

\paragraph{Instance Depth Loss.}
This loss is a L1 loss for instance depths:
\begin{align}
&\mathcal{L}_{zc} = \sum_{\mathbf{g}} \mathbb{1}_{\mathbf{g}}^{obj} \cdot d(Z_{cc}^{\mathbf{g}}, \widetilde{Z}_c^{\mathbf{g}}) \nonumber \\
&\mathcal{L}_{z\delta} = \sum_{\mathbf{g}} \mathbb{1}_{\mathbf{g}}^{obj}\cdot d(Z_{cc}^{\mathbf{g}} + \delta_{Z_c}^{\mathbf{g}}, \widetilde{Z}_c^{\mathbf{g}}) \nonumber \\
&\mathcal{L}_{depth} = \alpha \mathcal{L}_{zc} + \mathcal{L}_{z\delta}
\end{align}
where $\alpha > 1$ that encourages the network to first learn the coarse depths and then the deltas.

\paragraph{3D Localization Loss.}
This loss sums up the L1 loss of 2D projection and 3D location:
\begin{align}
&\mathcal{L}_{c}^{2d} =\sum_{\mathbf{g}} \mathbb{1}_{\mathbf{g}}^{obj} \cdot d( \mathbf{g} + \delta_{\mathbf{c}}^{\mathbf{g}}, \widetilde{\mathbf{c}}^{\mathbf{g}}) \nonumber\\
&\mathcal{L}_{c}^{3d} = \sum_{\mathbf{g}} \mathbb{1}_{\mathbf{g}}^{obj} \cdot d(\mathbf{C}_s^{\mathbf{g}} + \delta_{\mathbf{C}}^{\mathbf{g}}, \widetilde{\mathbf{C}}^{\mathbf{g}}) \nonumber\\
&\mathcal{L}_{location} = \beta \mathcal{L}_{c}^{2d} + \mathcal{L}_{c}^{3d}
\end{align}
where $\beta > 1$ to make it possible to learn the projected center first and then refine the final 3D prediction.

\paragraph{Local Corner Loss.}
The loss is the sum of L1 loss for all corners:
\begin{align}
\mathcal{L}_{corners} = \sum_{\mathbf{g}} \sum_k \mathbb{1}_{\mathbf{g}}^{obj}\cdot d(\mathbf{O}_k, \widetilde{\mathbf{O}}_k)
\end{align}

\subsubsection{Joint 3D Loss.}
Note that in the above loss functions , we decouple the monocular 3D detection into several subtasks and respectively regress different components of the 3D bounding box. Nevertheless, the prediction should be as a whole, and it is necessary to establish a certain relationship among different parts. 
We formulate the joint 3D loss as the sum of distances of corner coordinates in the camera coordinate frame:
\begin{align}
\mathcal{L}_{joint} = \sum_{\mathbf{g}} \sum_k \mathbb{1}_{\mathbf{g}}^{obj}\cdot d(\mathbf{O}_k^{cam}, \widetilde{\mathbf{O}}_k^{cam})
\end{align}

\subsection{Implementation Details}
\paragraph{Network Setup.}
The architecture of MonoGRNet is shown in \fig{\ref{fig:net}}. 
We choose VGG-16~\cite{matthew2014vgg} as the CNN backbone, but without its FC layers. We adopt KittiBox~\cite{teichmann2016multinet} for fast 2D detection and insert a buffer zone to separate 3D reasoning branches from the 2D detector. In the IDE module, a depth encoder structure similar in DORN~\cite{fu2018ordinal} is integrated to capture both local and global features. 
We present detailed settings for each layer in the supplemental material. 
There are 46 weighted layers in total, with only 20 weighted layers for the deepest path (\ie, from the input to the IDE output), due to the parallel 3D reasoning branches. In our design, there are 7.7 million parameters in all the 2D and 3D modules, which is approximately $6.2\%$ of the fully connected layers in the original VGG-16. 

\paragraph{Training.}
The VGG-16 backbone is initialized with the pretrained weights on ImageNet. In the loss functions, we set $\omega=\alpha=\beta=10$. L2 regularization is applied to the model parameters with a decay rate of 1e-5. We first train the 2D detector, along with the backbone, for 120K iterations using the Adam optimizer~\cite{kin2015adam}. Then the 3D reasoning modules, IDE, 3D localization and local corners, are trained for 80K iterations with the Adam optimizer. Finally, we use SGD to optimize the whole network in an end-to-end fashion for 40K iterations. The batch-size is set to 5, and the learning rate is 1e-5 throughout training. The network is trained using a single GPU of NVidia Tesla P40.
\section{Experiment}
\begin{figure}[t]
	\centering
	\includegraphics[width=1\linewidth]{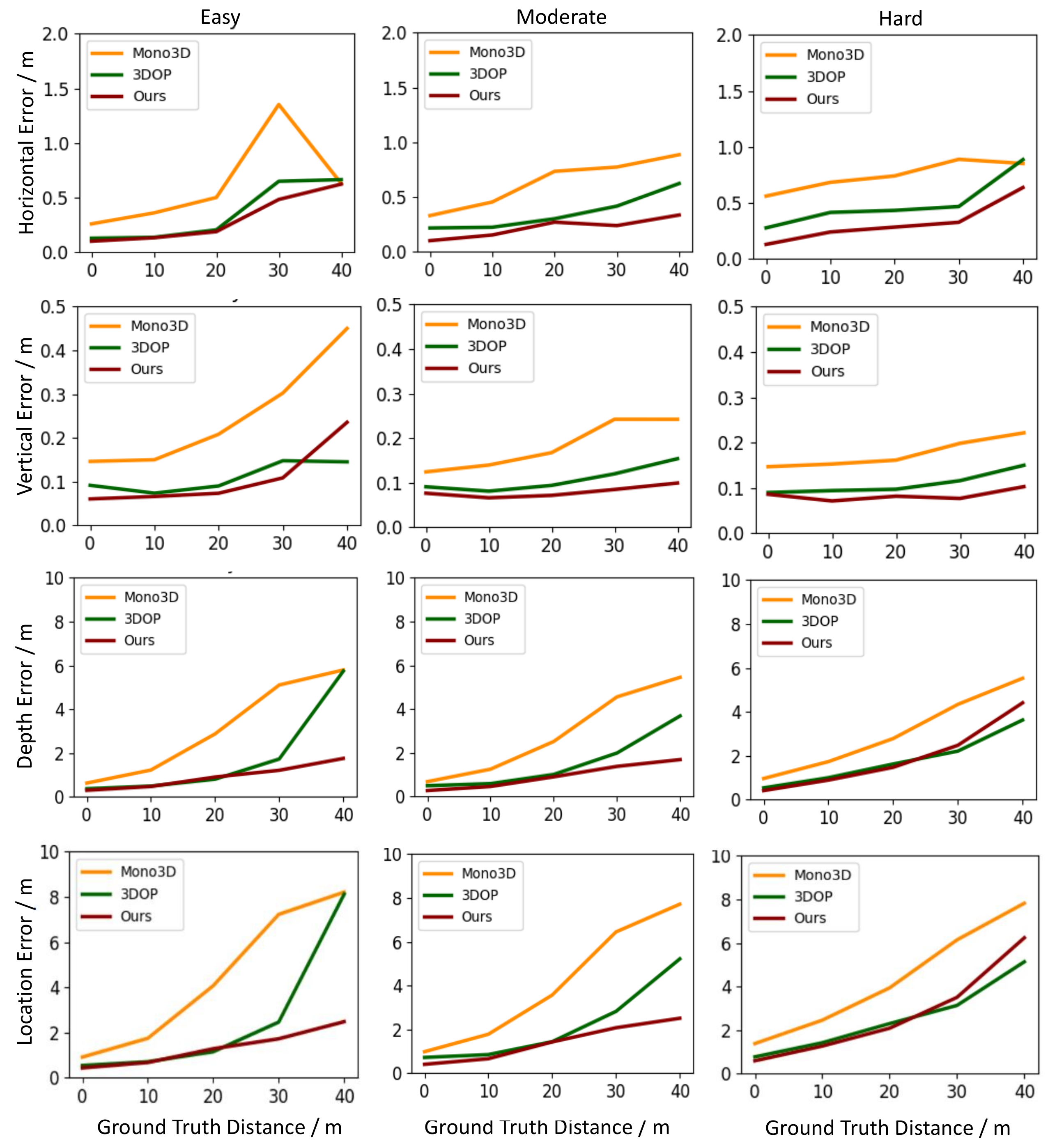}
	\caption{3D Localization Errors in the horizontal, vertical and depth dimensions according to the distances between objects and camera centers.}
	\label{fig:localization3d}
\end{figure}
\begin{table*}[t]   
	\centering
	\setlength{\tabcolsep}{2.1mm}{  
		\begin{spacing}{1.5}
			\scalebox{0.68}{
				\begin{tabular}{|p{2cm}<{\centering}|p{1cm}<{\centering}|c||ccc||ccc||ccc|}
					\hline
					\multirow{2}{*}{Method} & \multirow{2}{*}{Type} & \multirow{2}{*}{Time (s)}   & \multicolumn{3}{c||}{AP\textsubscript{3D} / AP\textsubscript{BEV} (IoU=0.3)} & \multicolumn{3}{c||}{AP\textsubscript{3D} / AP\textsubscript{BEV} (IoU=0.5)} & \multicolumn{3}{c|}{AP\textsubscript{3D} / AP\textsubscript{BEV} (IoU=0.7)}                  \\ \cline{4-12} 
					&                       &                       & \multicolumn{1}{c|}{Easy} & \multicolumn{1}{c|}{Moderate} & Hard              & \multicolumn{1}{c|}{Easy} & \multicolumn{1}{c|}{Moderate} & Hard  & \multicolumn{1}{c|}{Easy} & \multicolumn{1}{c|}{Moderate} & Hard           \\ \hline \hline
					3DOP                    & Stereo                &4.2  &69.79 / 71.41 &52.22 / 57.78 &\textbf{49.64} / \textbf{51.91} & 46.04 / \textbf{55.04}                     & 34.63 / \textbf{41.25}                & 30.09 / \textbf{34.55} & 6.55 / 12.63                     & 5.07 / 9.49                         & 4.10 / 7.59          \\      
					\hline       
					Mono3D                  & Mono                  & 3  &28.29 / 32.76 &23.21 / 25.15  &19.49 / 23.65 & 25.19 / 30.50                      & 18.20 / 22.39                        & 15.22 / 19.16         & 2.53 / 5.22                     & 2.31 / 5.19                         & 2.31 / 4.13          \\ 
					
					MF3D                    & Mono                  & 0.12 &/ &/ &/              & 47.88 / 55.02            & 29.48 / 36.73                        & 26.44 / 31.27         & 10.53 / 22.03                     & 5.69 / 13.63                         & 5.39 / 11.60          \\ 
					Ours                    & Mono                  & \textbf{0.06}  &\textbf{72.17} / \textbf{73.10} &\textbf{59.57} / \textbf{60.66} &46.08 / 46.86             &\textbf{50.51} / 54.21                   & \textbf{36.97} / 39.69                         & \textbf{30.82} / 33.06         & \textbf{13.88} / \textbf{24.97}           & \textbf{10.19} / \textbf{19.44}               & \textbf{7.62} / \textbf{16.30} \\ \hline
					
				\end{tabular}  
			} 
	\end{spacing}}
	\caption{\textbf{3D Detection Performance.} Average Precision of 3D bounding boxes on the same KITTI validation set and the inference time per image. Note that the stereo-based method 3DOP is not compared but listed for reference.
	}
	\label{tab:3dap} 
\end{table*}
We evaluate the proposed network on the challenging KITTI dataset~\cite{geiger2012kitti}, which contains 7481 training images and 7518 testing images with calibrated camera parameters. Detection is evaluated in three regimes: easy, moderate and hard, according to the occlusion and truncation levels. 
We compare our method with the state-of-art monocular 3D detectors, MF3D ~\cite{xu2018multifusion} and Mono3D~\cite{chen2016monocular}. We also present the results of a stereo-based 3D detector 3DOP~\cite{chen20153dop} for reference. For a fair comparison, we use the train1/val1 split following the setup in~\cite{chen2016monocular,chen2017multiview}, where each set contains half of the images.

\paragraph{Metrics.} 
For evaluating 3D localization performance, we use the mean errors between the central location of predicted 3D bounding boxes and their nearest ground truths.
For 3D detection performance, we follow the official settings of KITTI benchmark to evaluate the 3D Average Precision (AP\textsubscript{3D}) at different Intersection of Union (IoU) thresholds.  

\paragraph{3D Localization Estimation.}
We evaluate the three dimensional location errors (horizontal, vertical and depth) according to the distances between the targeting objects and the camera centers. The distances are divided into intervals of 10 meters. The errors are calculated as the mean differences between the predicted 3D locations and their nearest ground truths in meters. Results are presented in \fig{\ref{fig:localization3d}}. The errors, especially for in the depth dimension, increase as the distances grow because far objects presenting small scales are more difficult to learn.

The results indicate that our approach (red curve) outperforms Mono3D by a significant margin, and is also superior to 3DOP, which requires stereo images as input. Another finding is that, in general, our model is less sensitive to the distances. When the targets are 30 meters or farther away from the camera, our performance is the most stable, indicating that our network deals with far objects(containing small image regions) best.

Interestingly, horizontal and vertical errors are an order of magnitude smaller than that of depth, \ie, the depth error dominants the overall localization error. This is reasonable because the depth dimension is not directly observed in the 2D image but is reasoned from geometric features. The proposed IDE module performs superior to the others for the easy and moderate regimes and comparable to the stereo-based method for the hard regime. 

%
%
%

\paragraph{3D Object Detection.}
3D detection is evaluated using the AP\textsubscript{3D} at 0.3, 0.5 and 0.7 3D IoU thresholds for the car class. We compare the performance with two monocular methods, Mono3D and MF3D. The results are reported in \tab{\ref{tab:3dap}}. Since the authors of MF3D have not published their validation results, we only report the AP\textsubscript{3D} at 0.5 and 0.7 presented in their paper. Experiments show that our method outperforms the state-of-art monocular detectors mostly and is comparable to the stereo-based method.

Our network is designed for efficient applications and a fast 2D detector with no region proposal is adopted. The inference time achieves about 0.06 seconds per image on a Geforce GTX Titan X, which is much less than the other three methods. On the other hand, this design at some degree sacrifices the accuracy of 2D detection. Our 2D AP of the moderate regime at 0.7 IoU threshold is 78.14\%, about 10\% lower than the region proposal based methods that generate a large number of object proposals to recall as many groundtruth as possible. Despite using a relatively weak 2D detector, our 3D detection achieves the state-of-the-art performance, resorting to our IDE and 3D localization module. Note that the 2D detection is a replaceable submodule in our network, and is not our main contribution. 

\begin{figure*}[t] 
	\centering
	\scriptsize
	\begin{tabular}{c@{\hspace{0.14cm}}c@{\hspace{0.14cm}}c}
		\includegraphics[width=0.32\linewidth]{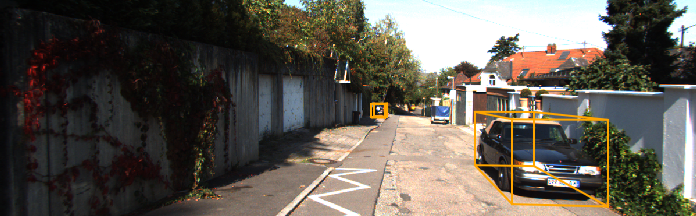} &
		\includegraphics[width=0.32\linewidth]{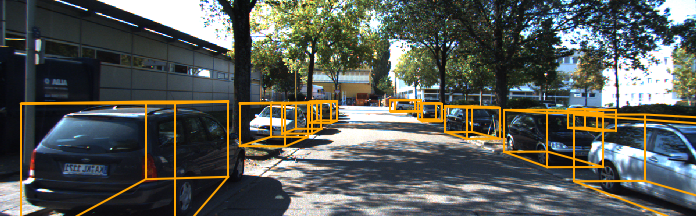} &
		\includegraphics[width=0.32\linewidth]{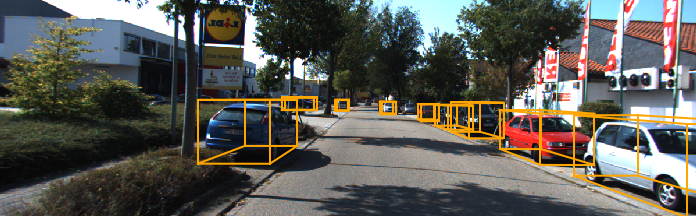} \\
		\includegraphics[width=0.32\linewidth,trim={0cm 6cm 0cm 1cm},clip]{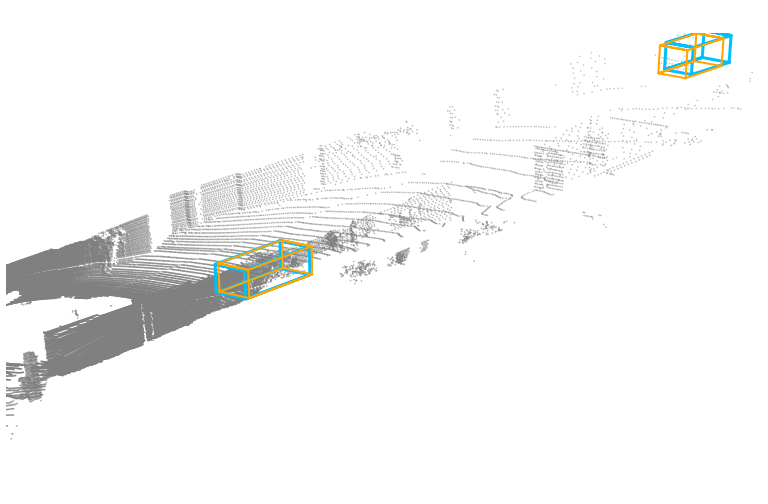} &
		\includegraphics[width=0.32\linewidth,trim={0cm 6cm 0cm 1cm},clip]{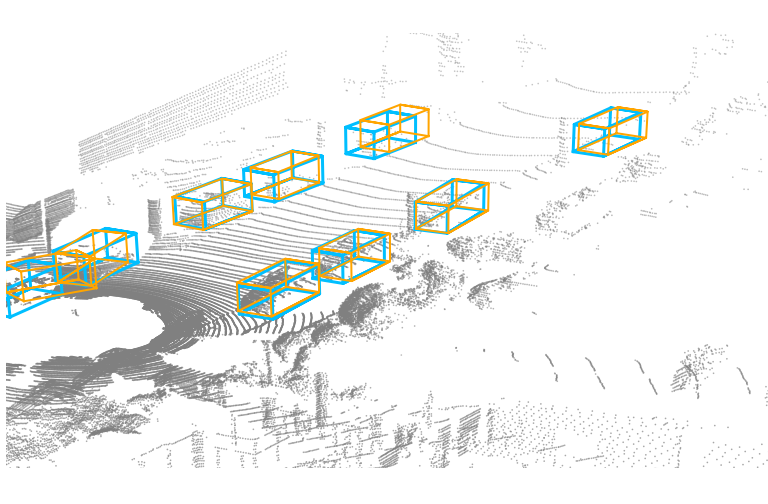} &
		\includegraphics[width=0.32\linewidth,trim={0cm 6cm 0cm 1cm},clip]{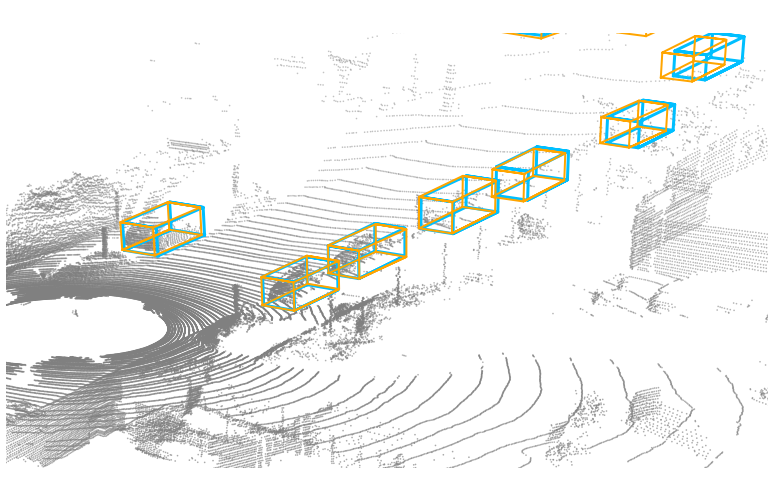} \\
		(a)  & (b)  & (c) \\
		
		\includegraphics[width=0.32\linewidth]{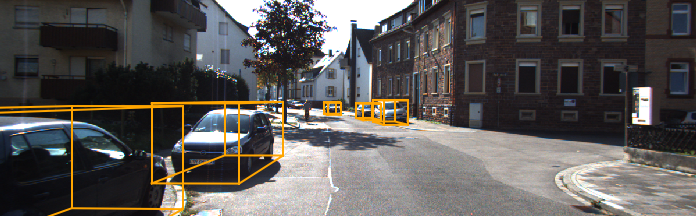} &
		\includegraphics[width=0.32\linewidth]{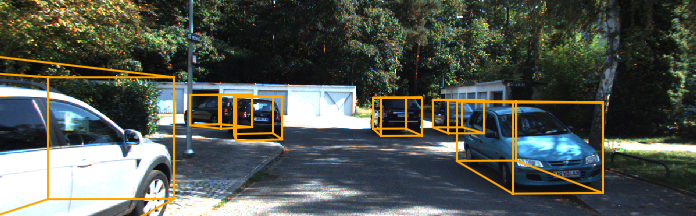} &
		\includegraphics[width=0.32\linewidth]{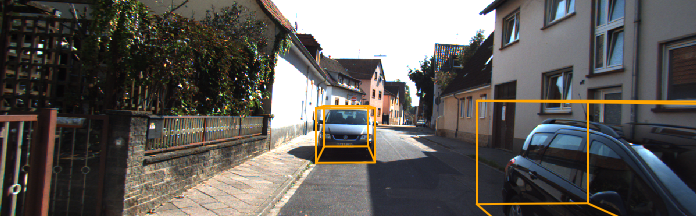} \\
		\includegraphics[width=0.32\linewidth,trim={0cm 6cm 0cm 1cm},clip]{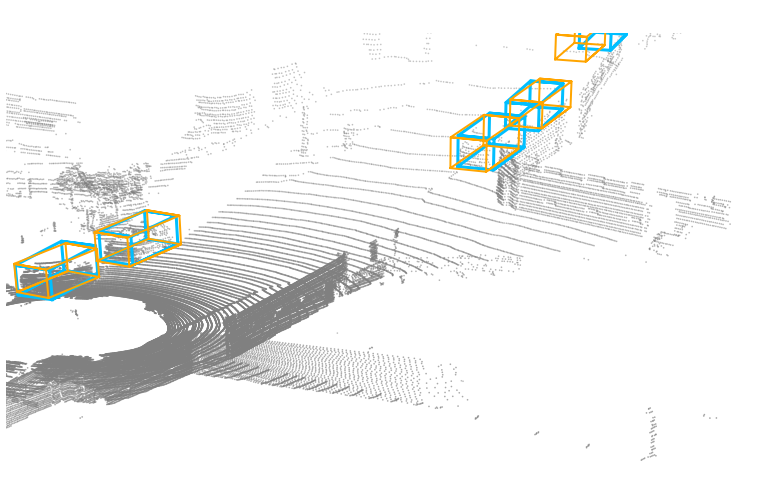} &
		\includegraphics[width=0.32\linewidth,trim={0cm 6cm 0cm 1cm},clip]{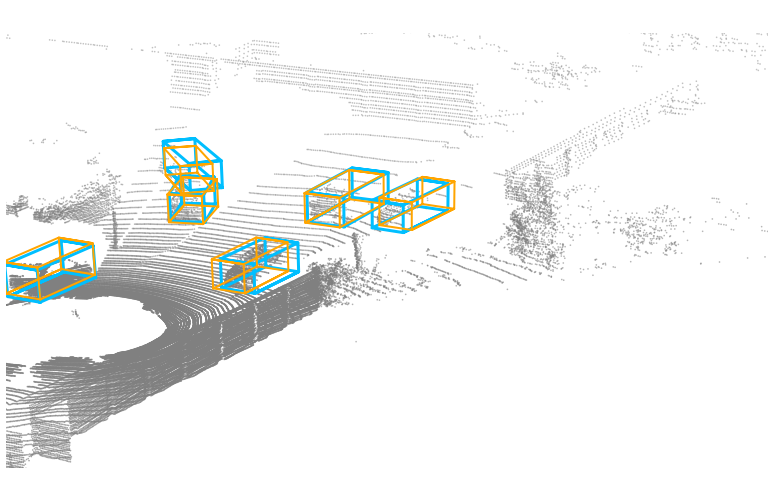} &
		\includegraphics[width=0.32\linewidth,trim={0cm 6cm 0cm 1cm},clip]{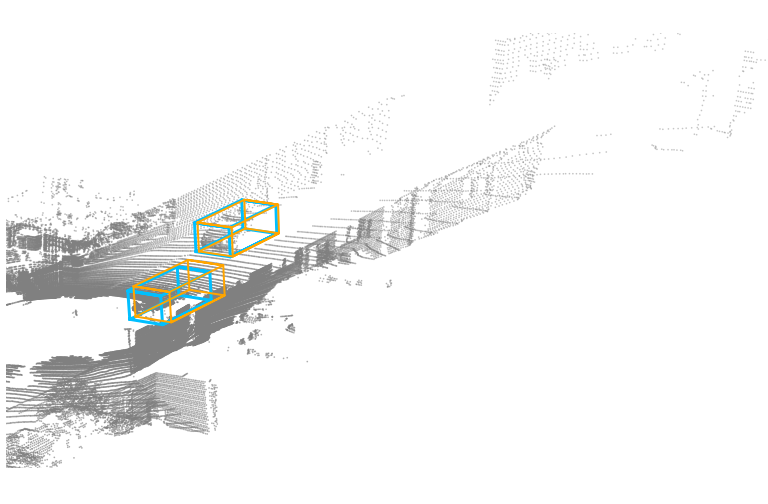} \\
		(d)  & (e)  & (f) \\
		
		\includegraphics[width=0.32\linewidth]{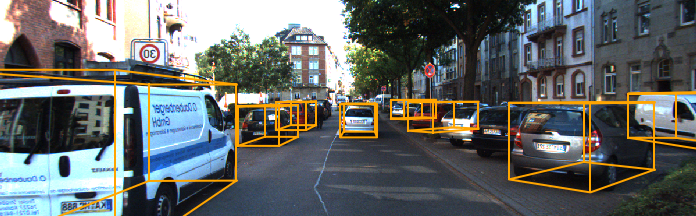} &
		\includegraphics[width=0.32\linewidth]{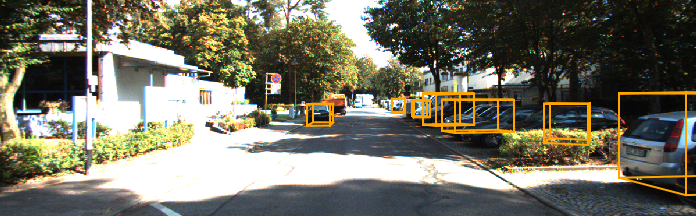} &
		\includegraphics[width=0.32\linewidth]{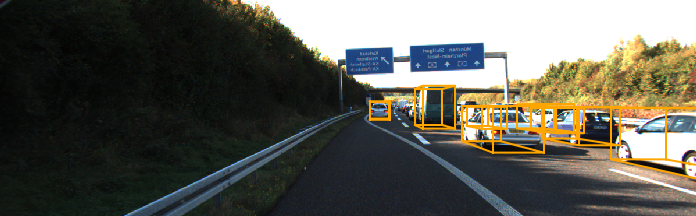} \\
		\includegraphics[width=0.32\linewidth,trim={0cm 6cm 0cm 1cm},clip]{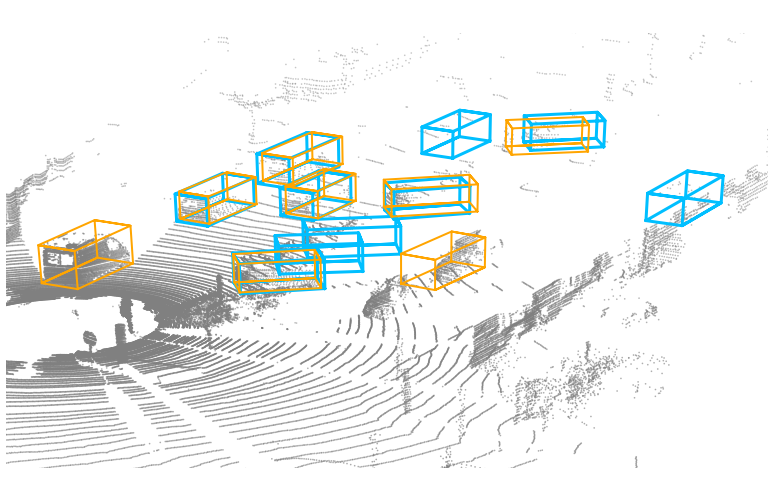} &
		\includegraphics[width=0.32\linewidth,trim={0cm 6cm 0cm 1cm},clip]{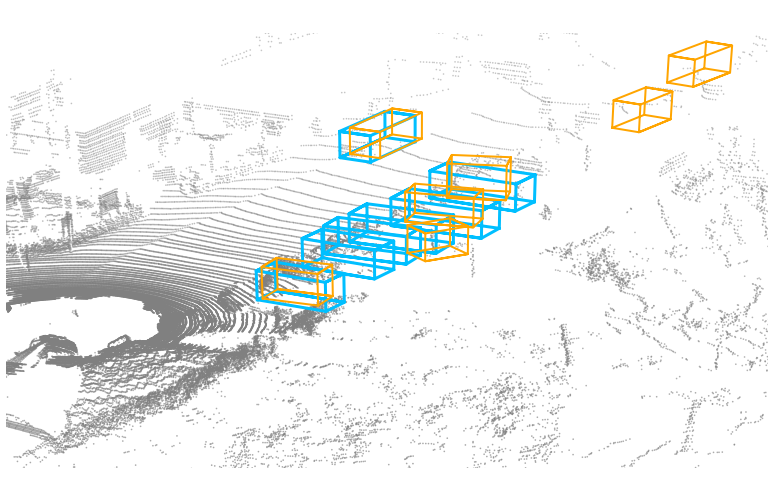} &
		\includegraphics[width=0.32\linewidth,trim={0cm 6cm 0cm 1cm},clip]{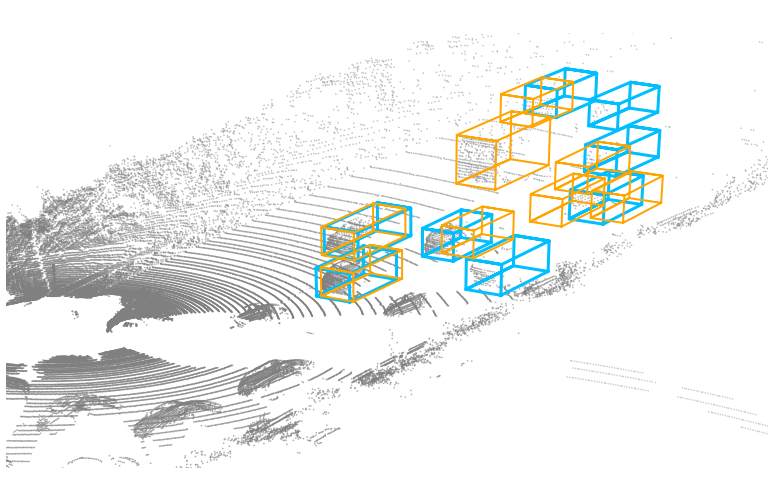} \\
		(g)  & (h)  & (i)
		
	\end{tabular}
	\caption{\textbf{Qualitative Results.} Predicted 3D bounding boxes are drawn in orange, while ground truths are in blue. Lidar point clouds are plotted for reference but not used in our method. Camera centers are at the bottom-left corner. (a), (b) and (c) are common cases when predictions recall the ground truths, while (d), (e) and (f) demonstrate the capability of our model handling truncated objects outside the image. (g), (h) and (i) show the failed detections when some cars are heavily occluded.}
	\label{fig:visualize3d}
\end{figure*}

\paragraph{Local 3D Bounding Box Regression.}
We evaluate the regression of local 3D bounding boxes with the size (height, width, length) and orientation metrics.
The height, width, length of a 3D bounding box can be easily calculated from its eight corners. The orientation is measured by the azimuth angles in the camera coordinate frame. We present the mean errors in \tab{\ref{tab:3dparams}}. Our network demonstrates a better capability to learn the size and orientation of a 3D bounding box from merely photometric features. 
It is worth noting that in our local corner regression module, after RoiAlign layers, all the objects of interest are rescaled to the same size to introduce scale-invariance, yet the network still manages to learn their real 3D sizes. 
This is because our network explores the image features that convey projective geometries and semantic information including types of objects (\eg, SUVs are generally larger than cars) to facilitate the size and orientation estimation.

\paragraph{Qualitative Results.}
 Qualitative visualization is provided for three typical situations, shown in \fig{\ref{fig:visualize3d}}. In common street scenes, our predictions are able to successfully recall the targets. It can be observed that even though the vehicles are heavily truncated by image boundaries, our network still outputs precise 3D bounding boxes. Robustness to such a corner case is important in the scenario of autonomous driving to avoid collision with lateral objects. For cases where some vehicles are heavily occluded by others, \ie, in (g), (h) and (i), our 3D detector can handle those visible vehicles but fails in detecting invisible ones. In fact, this is a general limitation of perception from monocular RGB images, which can be solved by incorporating 3D data or multi-view data to obtain informative 3D geometric details.

\begin{table}[]
		\setlength{\tabcolsep}{3mm}{
		\begin{spacing}{1.3}
\scalebox{0.88}{
	\begin{tabular}{|cccc|c|}
		\hline
		\multicolumn{1}{|c||}{\multirow{2}{*}{Method}} & \multicolumn{3}{c|}{Size (m)}                                                                          & \multirow{2}{*}{Orientation (rad)} \\ \cline{2-4}
		\multicolumn{1}{|c||}{}                        & Height                           & Width                            & Length                           &                                    \\ \hline \hline
		\multicolumn{1}{|c||}{3DOP}                  & 0.107                           & 0.094                           & 0.504                           & 0.580                            \\
		\multicolumn{1}{|c||}{Mono3D}                    & 0.172                          & 0.103                          & 0.582                          &  0.558                            \\
		\multicolumn{1}{|c||}{Ours}                    &\textbf{0.084}  &\textbf{0.084}  &\textbf{0.412}  & \textbf{0.251}   \\ \hline
	\end{tabular}
}
\end{spacing}}
\caption{{3D Bounding Box Parameters Error.}} 
\label{tab:3dparams}
\end{table}

%
%

\subsection{Ablation Study}
A crucial step for localizing 3D center $\mathbf{C}$ is estimating its 2D projection $\mathbf{c}$, since $\mathbf{c}$ is analytically related to $\mathbf{C}$. Although the 2D bounding box's center $\mathbf{b}$ can be close to $\mathbf{c}$, as is illustrated in \fig{\ref{fig:notation}} (a), it does not have such 3D significance. When we replace $\mathbf{c}$ with $\mathbf{b}$ in 3D reasoning, the horizontal location error rises from 0.27m to 0.35m, while the vertical error increases from 0.09m to 0.69m. Moreover, when an object is truncated by the image boundaries, its projection $\mathbf{c}$ can be outside the image, while $\mathbf{b}$ is always inside. In this case, using $\mathbf{b}$ for 3D localization can result in a severe discrepancy. Therefore, our subnetwork for locating the projected 3D center is indispensable.


In order to examine the effect of coordinate transformation before local corner regression, we directly regress the corners offset in camera coordinates without rotating the axes. It shows that the average orientation error increases from 0.251 to 0.442 radians, while the height, width and length errors of the 3D bounding box almost remain the same. This phenomenon corresponds to our analysis that switching to object coordinates can reduce the rotation ambiguity caused by projection, and thus enables more accurate 3D bounding box estimation.


\section{Conclusion}
We have presented the MonoGRNet for 3D object localization from a monocular image, which achieves superior performance on 3D detection, localization and pose estimation among the state-of-the-art monocular methods. A novel IDE module is proposed to predict precise instance-level depth, avoiding extra computation for pixel-level depth estimation and center localization, regardless of far distances between objects and camera.
Meanwhile, we distinguish the 2D bounding box center and the projection of 3D center for a better geometric reasoning in the 3D localization. The object pose is estimated by regressing corner coordinates in a local coordinate frame that alleviates ambiguities of 3D rotations in perspective transformations. The final unified network integrates all components and performs inference efficiently.

\par\vfill\par
\clearpage

\bibliographystyle{aaai}
\bibliography{reference_mod}

\clearpage

\end{document}